\documentclass[10pt,twocolumn,letterpaper]{article}

\usepackage{iccv}
\usepackage{times}
\usepackage{epsfig}
\usepackage{graphicx}
\usepackage{amsmath}
\usepackage{amssymb}

\usepackage[ruled]{algorithm}
\usepackage{algpseudocode}

\newcommand{\keypoint}[1]{\vspace{0.0cm}\noindent\textbf{#1}\quad}	

\usepackage[breaklinks=true,bookmarks=false]{hyperref}

\iccvfinalcopy 

\def\iccvPaperID{2421} 

\ificcvfinal\fi
\begin{document}

\title{Goal-Driven Sequential Data Abstraction}


\author{Umar Riaz Muhammad$^1$ \quad Yongxin Yang$^1$ \quad Timothy M. Hospedales$^{2}$ \quad Tao Xiang$^1$ \quad Yi-Zhe Song$^1$\\
    $^1$University of Surrey \quad \quad
    $^2$University of Edinburgh \\
    {\tt\small \{u.muhammad, yongxin.yang, t.xiang, y.song\}@surrey.ac.uk, t.hospedales@ed.ac.uk}
}

\maketitle
\ificcvfinal\thispagestyle{empty}\fi

\begin{abstract}
    Automatic data abstraction is an important capability for both benchmarking machine intelligence and supporting summarization applications. In the former one asks whether a machine can `understand' enough about the meaning of input data to produce a meaningful but more compact abstraction. In the latter this capability is exploited for saving space or human time by summarizing the essence of input data. 
    In this paper we study a general reinforcement learning based framework for learning to abstract sequential data in a goal-driven way. The ability to define different abstraction goals uniquely allows different aspects of the input data to be preserved according to the ultimate purpose of the  abstraction. Our reinforcement learning objective does not require human-defined examples of ideal abstraction. Importantly our model processes the input sequence holistically without being constrained by the original input order. Our framework is also domain agnostic -- we demonstrate applications to sketch, video and text data and achieve promising results in all domains. 
\end{abstract}

\section{Introduction}

Abstraction is generally defined in the context of specific applications  \cite{bransford1971abstraction, kang2009flow, truong2007video, cheng2013efficient, mani1999advances, muhammad2018learning}. In most cases it refers to elimination of redundant elements, and preservation of the most salient and important aspects of the data. It is an important capability for various reasons: compression \cite{gandomi2015beyond} and saving human time in viewing the data \cite{narayan2018ranking}; but also improving downstream data analysis tasks such as information retrieval \cite{aliguliyev2009new}, and synthesis \cite{graves2013generating, muhammad2018learning}.

\begin{figure}
    \centering
	\includegraphics[width=\linewidth]{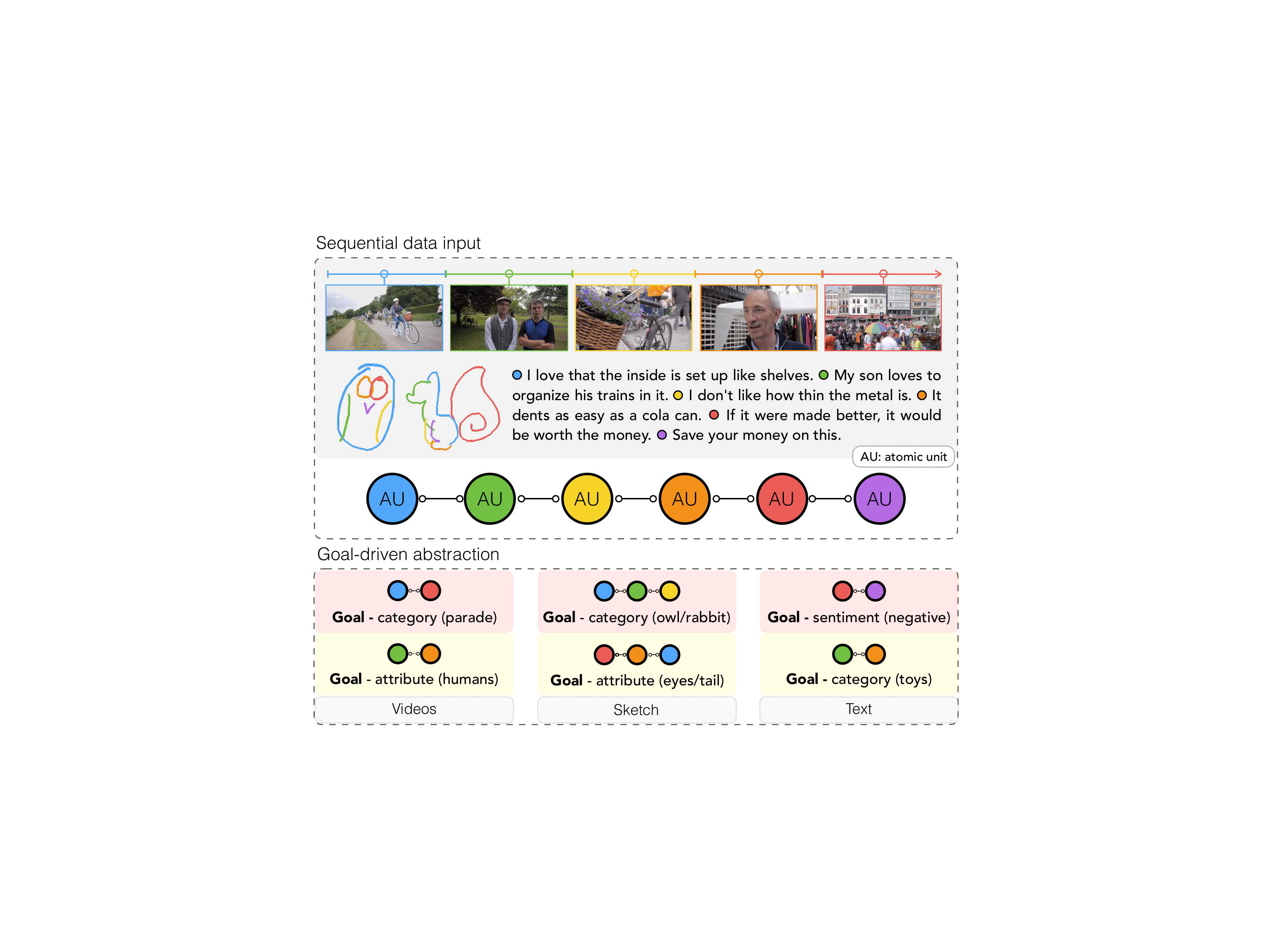}
	\caption{An illustration of our goal-driven abstraction task. Each input (video, sketch or text) consists of a sequence of atomic units (AUs) corresponding to  video-segments, strokes and sentences for the three input domains, respectively. The AUs are color coded.  Depending on the abstraction goal, different AUs are preserved in each abstracted output.}
	\vspace{-0.5cm}
	\label{fig:Preview}
\end{figure} 

We present a novel goal-driven abstraction task for sequential data (see Fig.~\ref{fig:Preview}). Sequential refers to data with temporal order -- we consider video, sequentially drawn sketches, and text. Goal-driven refers to preservation of a certain aspect of the input according to a specific abstraction objective or goal. The same input may lead to different abstracted outputs depending on the abstraction goal. For example, prioritizing preserving sentiment vs.~helpfulness in product review text could lead to different summaries. 
It is important not to confuse our new goal-driven abstraction setting with  traditional video/text summarization  \cite{ gygli2014creating, gygli2015video, wei2018video, zhao2018hsa, sutskever2014sequence, cheng2016neural, nallapati2017summarunner, narayan2017neural, yasunaga2017graph}. The aims are different: the latter produces a \emph{single} compact but diverse and representative summary, often guided by human annotations, while ours yields \emph{various} goal-conditional compact summaries. Our problem setting is also more amenable to training without ground-truth labels (i.e., manual gold-standard but subjective summaries) commonly required by contemporary video/text summarization methods.

To tackle this novel problem, new approaches are needed. To this end, we propose a goal-driven sequential data abstraction model with the following key properties: (1) It processes the input sequence holistically rather than being constrained by the original input order. (2) It is trained by reinforcement learning (RL), rather than supervised learning. This means that expensively annotated data in the form of target abstractions are not required.
(3) Different goals are introduced via RL reward functions. Besides eliminating the annotation requirement, this enables preserving different aspects of the input according to the  purpose of the abstraction. (4) Finally, the RL-based approach also allows abstracted outputs of any desired length to be composed by varying the abstraction budget.

We demonstrate the generality of our approach through three very different sequential data domains: free-hand sketches, videos and text. Video and text are sequential data domains widely studied in the past. While sketch may not seem obviously sequential, touchscreen technology means that all prominent sketch datasets now record vectorized stroke sequences. For instance, QuickDraw \cite{ha2017neural}, the largest sketch dataset to date, provides vectorized sequence data in the form of $(x,y)$ pen coordinates and state $p$ (touching or lifting). For sketch and video, we train two reward functions based on \emph{category} and \emph{attribute} recognition models. These drive our abstraction model to abstract an input sketch/video into a shorter sequence, while selectively preserving either category or attribute related information. For text, we train three reward functions on product reviews, based on  \emph{sentiment}, \textit{product-category} and \textit{helpfulness} recognition models. These drive our model to summarize an input document into a shorter paragraph that preserves sentiment/category/helpfulness information respectively.

The main contributions of our work are: (1) Defining a novel goal-driven abstraction problem, (2) a sequential data abstraction model trained by RL, that processes the input holistically without being constrained by the original input order, and  (3) demonstrating flexibility of this model to diverse sequential data domains including sketch, video and text.

\section{Related work}
\label{sec:RelatedWork}
\keypoint{Video/text summarization} Existing models are either supervised or unsupervised. Unsupervised summarization models in video \cite{elhamifar2012see, otani2016video, wang2016video, wang2017discovering, yang2015unsupervised, zhang2018dtr, zhao2014quasi, zhu2013video, zhu2016learning} and text  \cite{erkan2004lexrank,mihalcea2004textrank, mcdonald2007study, baralis2013graphsum} domains aim to  identify a small subset of key  units (video-segments/sentences) that preserve the global content of the input, e.g., using criteria like diversity and representativeness. In contrast, supervised video \cite{gong2014diverse, gygli2014creating, gygli2015video, wei2018video, zhang2016video, zhao2018hsa} and text \cite{sutskever2014sequence, cheng2016neural, nallapati2017summarunner, narayan2017neural, yasunaga2017graph} summarization methods solve the same problem by employing ground-truth summaries as training targets. Both types of models are not driven by  specific goals and are evaluated on human annotated ground-truth summaries -- how humans summarize a given video/text is subjective and often ambiguous.   Neither of these models thus address our new goal-driven abstraction setting.

A recent work \cite{zhou2018video} trains video summarization model in a weakly supervised RL setting using category level video labels. The aim is to produce summaries with the added criteria of category level recognizability along with the usual criteria of diversity and representativeness. The core mechanism is to process  video-segments in sequence and make binary decisions (keep or remove) for each segment, following the above criteria. In this work we introduce a goal-driven approach to explicitly preserve any quantifiable property, whether category information (as partially done in \cite{zhou2018video}), attributes, or potentially other quantities such as interestingness \cite{fu2014ranking}.  We show that our model is superior to \cite{zhou2018video} thanks to the holistic modeling of the sequential input without restriction by its original order (see Sec.~\ref{subsec:VideoAbstraction}). 

\keypoint{Sketch abstraction} Comparing to video and text, much less prior work on sketch abstraction exists. This problem was first studied in \cite{berger2013style} where a data-driven approach was used to study abstraction in professionally drawn facial portraits. Sketches at varying abstraction levels were collected by limiting the time (from four and a half minutes to fifteen seconds) given to an artist to sketch a reference photo. In recent work \cite{muhammad2018learning}, automatic abstraction was studied  explicitly for the first time in free-hand amateur sketches. The abstraction process was defined as a trade-off between recognizability and brevity/compactness of the sketch. The abstraction model, also based on RL,  processed stroke-segments in sequence and made binary decisions (keep or remove) for each segment, but otherwise output strokes in the same order as they were  drawn. In this work we also optimize the trade-off between recognizability and compactness (if the goal is recognizability). However, crucially, our method benefits from processing the input holistically rather than in its original order, and learns an optimal stroke sequencing strategy.  We show that our approach clearly outperforms \cite{muhammad2018learning} (see Sec.~\ref{subsec:SketchAbstraction}). Further, we demonstrate application to diverse domains of sketch, video and text, and  uniquely explore the ability to use multiple goal functions to obtain different abstractions.

\keypoint{Sketch recognition} Early sketch recognition methods were developed to deal with professionally drawn sketches as in CAD or artistic drawings \cite{jabal2009comparative, lu2005new, sousa2009geometric}. The more challenging task of free-hand sketch recognition was first tackled in \cite{eitz2012humans} along with the release of the first large-scale dataset of amateur sketches. Since then the task has been well-studied using both classic vision   \cite{schneider2014sketch, li2015free} as well as deep learning approaches  \cite{yu2015sketch}. Recent successful deep learning approaches have spanned both primarily non-sequential CNN \cite{yu2015sketch,yu2017sketch} and sequential RNN  \cite{jia2017sequential,sarvadevabhatla2018game} recognizers. We use both CNN and RNN-based multi-class classifiers to provide rewards for our RL based sketch abstraction framework.

\section{Methodology}
\label{sec:Methodology}

\begin{figure*}[h]
	\includegraphics[width=\textwidth]{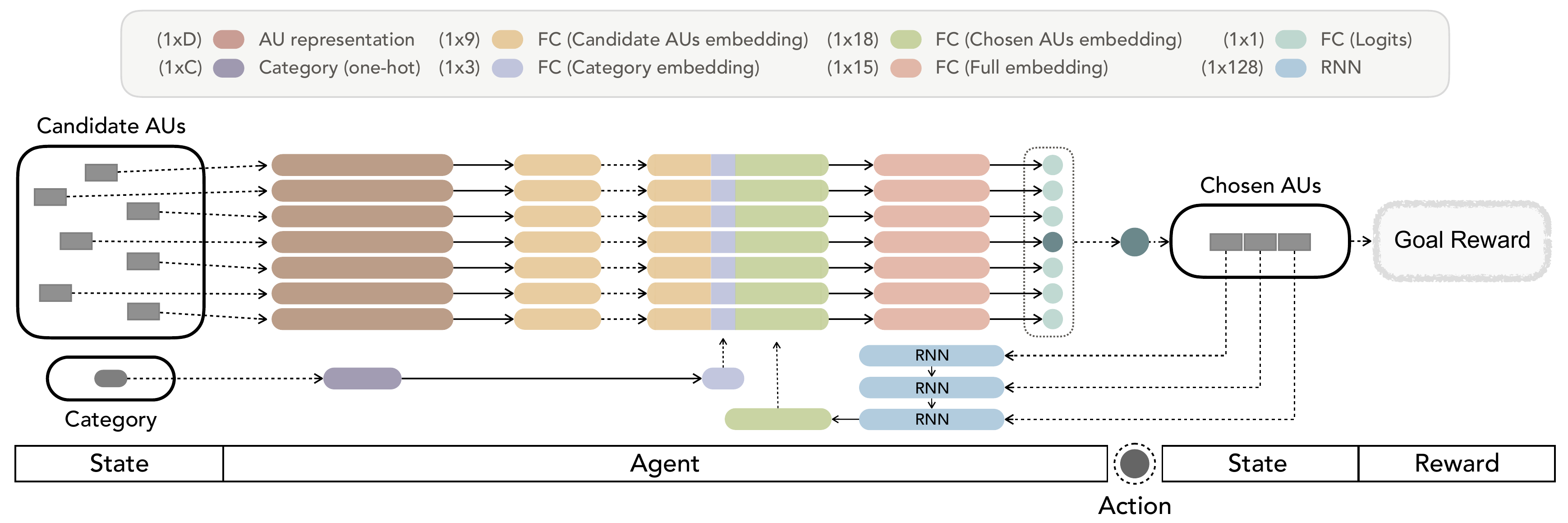}
	\caption{A schematic illustration of the proposed GDSA agent. The agent iteratively chooses AUs from the candidate pool so as to maximize the recognizability goal of the abstracted sketch/text/video. Solid arrows represent trainable weights.}
	\vspace{-0.5cm}
	\label{fig:Model}
\end{figure*}

Our aim is to input a data sequence and output a shorter sequence that preserves a specific type of information according to a goal function. To this end, a goal-driven sequence abstraction (GDSA) model is proposed. GDSA processes the input sequence data holistically by first decomposing it into a set of atomic-units (AUs), which form a pool of candidates for selection. GDSA is trained by RL to produce abstractions by picking a sequence of AUs from the pool. The output sequence should be shorter than the input (controlled by the budget) while preserving its information content (controlled by the RL reward/goal function).

\subsection{Goal-driven sequence abstraction (GDSA)}
\label{subsec:GDSAmodel}

The sequential data abstraction task is formalized as a Markov decision process. At each step, our GDSA agent moves one atomic unit (AU) from a pool of candidate AUs to a list of chosen AUs, and it stops when the number of chosen AUs is larger than a fixed budget. The agent is trained via RL \cite{Sutton98Reinforcement} using a reward scheme which encourages it to outperform the efficiency of the original input order in terms of preserving goal-related information in the sequence given a limited length budget. Concretely, we have two data-structures: the candidate AU pool and the chosen AU list. The chosen AUs list starts empty, and the candidate AUs pool contains the full input. AUs are then picked by the agent from the candidate pool, one at a time, to be appended to the current chosen AU list.

A schematic of the GDSA agent is shown in Fig.~\ref{fig:Model}. The core idea is to evaluate the choice of each candidate AU in the context of all previously chosen AUs and the category to which the input sequence belongs. We do this by learning embeddings for candidate AUs, chosen AUs, and the input sequence category label respectively. Based on these embeddings, the GDSA agent will iteratively pick the next best AU to output given those chosen so far.

\keypoint{Candidate AU embedding} At each iteration, every AU in the candidate pool is considered by GDSA as a candidate for the next output. To this end, first each AU is: (1) Encoded as a fixed-length vector. Note that each AU may itself contain sequential sub-structure (sketch strokes formed by segments, video segments formed by frames, or sentences formed by words), so we use a domain-specific pre-trained RNN to embed each AU. The hidden RNN cell state corresponding to the last sub-entry of the AU is extracted and used to represent the AU as a fixed-length vector. (2) Assigned a time-stamp from 1 to 10 based on the relative position in the original input sequence w.r.t the total number of AUs. This is introduced so that during training our model can leverage information from the input sequence order. This one-hot time-stamp vector is then concatenated with the fixed-length RNN encoding vector above, and these are fed into a fully-connected (FC) layer to get the candidate AU embedding.

\keypoint{Chosen AU embedding} To represent the output sequence so far, all the AUs of the chosen AU list are fed sequentially to an RNN. Each AU corresponds to a time-step in the RNN. The output of the last time-step is then fed into a FC layer to get the chosen-AU list embedding. At the first time step the list is empty, represented by a zero vector.

\keypoint{Category embedding} There are often multiple related abstraction tasks within a domain, e.g., the object/document category in sketch/text abstraction. We could train an independent GDSA model per category, or aggregate the training data across all categories. These suffer respectively from less training data, and a mixing of category/domain-specific nuances. As a compromise we embed a category identifier to allow the model to exploit some information sharing, while also providing guidance about category differences \cite{yang2015mdmtl}.

\keypoint{Action selection} At each iteration, our agent takes an action (pick an AU from the candidate pool) given the category and chosen AUs so far. To this end, it considers each candidate AU in turn and concatenates it with the other two embeddings, before feeding the result into a FC layer to get a complete state-action embedding. This is then fed into a FC layer with $1$ neuron (i.e. scalar output) to produce the final logit. Once all candidate AUs are processed, their corresponding logit values are concatenated and form a multinomial distribution via softmax. During training, we sample this multinomial, and during testing the largest logit  is always chosen. The picked AU is then removed from the candidate pool and appended to the list of chosen AUs. This process is repeated until a budget is exhausted.

\keypoint{Domain specific details} We apply our framework to sketch, video and text data. Each sketch is composed of a sequence of AUs corresponding to strokes.
For video, each input is a video-clip and segments in the clip are AUs. For text, each input is a document containing a product review, and sentences are AUs.
Another domain specific property is on how to present the agent-picked AUs as the final output of the abstraction. In the case of video and text, the selected  AUs  are kept in the same order as in the original input order, to maintain the coherence of output sequence. While for sketch, we keep the order in which AUs are picked, since the model can potentially learn a better sequencing strategy than the natural human input.

\subsection{Goal-driven reward function}
\label{subsec:GoalDrivenRewardFunction}

The objective of our GDSA agent is to choose AUs that maximally preserve the goal information. In particular, we leverage the natural input sequence, along with the random AU selection, to define a novel reward function $r_t$ as 
\begin{equation}
\label{eq:reward}
r_{t} = ( a_{t} - (\delta  \: h_{t}  + (1 - \delta) \:  g_{t}) ) b,
\end{equation}
\noindent where $t$ is the time step\footnote{A time step means a pass through the candidate AU pool, leading to the selection of a chosen AU.}, $a_{t}$ is agent performance (at goal information preservation), $h_{t}$ is the performance obtained by picking the AUs following the original input order, and $g_{t}$ is the performance of random order policy. The performance is evaluated according to the recognizability of the goal information to be preserved after adding the selected AU to the chosen AUs list. $\delta$ is an annealing parameter that balances comparison against human and random-policy performance. It is initialised to $0$, so the agent receives positive reward as long as it beats the random policy. During training $\delta$ is increased  towards $1$, thus defining a curriculum that progressively requires the agent to perform better in order to obtain a reward. In detail, $\delta$ is increased from $0$ to $1$ linearly in $K$ steps, where $K$ is the total number of episodes used during training. So by the end of training, the agent's selection has to beat the original input sequence to obtain positive reward. Finally, $b$ is a reward scaling factor. For example, given a 100-stroke sketch and budget of 10\%, GDSA has to pick 10 strokes more informative than a random selection to obtain reward at the start, and more informative than the first 10 strokes of the input to obtain reward at the end.

\keypoint{Goals} 
For sketches, one abstraction goal is the recognizability of the output sequential object sketch. We quantify this by the resulting classification accuracy under a multi-class classifier (thus defining $a_t,h_t$ and $g_t$ in the reward function). To demonstrate driving abstraction by different goals, we explore rewarding preservation of other information about the sketch. Specifically, we train a sketch attribute detector to define an attribute preservation reward.
For videos, the main target information to be preserved is the recognizability of the video category. To guide training we employ a multi-class classifier, which is plugged into the reward function to compute $a_t,h_t$ and $g_t$ values at each time step. We also consider another  abstraction goal of preserving attributes in videos by employing an attribute detector to define the reward. 
For text, the main goal is sentiment preservation in product reviews, and the reward is given by probability of the review summary being correctly classified by a binary sentiment classifier. As different abstraction goals, we also explore the preservation of product-category and helpfulness information by training separate classifiers for these goals.

\subsection{Training procedure}
\label{subsec:TrainingProcedure}

\keypoint{Variable action space}
In a conventional reinforcement learning (RL) framework, the observation and action space dimensions are both fixed. In our framework, because the number of candidate AUs is decremented at each step, the action space shrinks over time. In contrast, the number of chosen AUs increases over time, but their embedding dimension is fixed, due to the use of RNN embedding. Our RL framework deals with these dynamics by rebuilding the action space at every time-step. This can be efficiently implemented by convolution over available actions (i.e., the candidate AU pool).

\keypoint{Objective}
The objective of RL is to find the policy $\pi$ that produces a trajectory $\tau$ of states and actions leading to the maximum expected reward. In our context the trajectory is the sequence of extracted AUs. The policy is realized by a neural network parametrized by $\theta$, i.e., $\pi_\theta$, where $\theta$ is the set of parameters for all modules mentioned in Sec.~\ref{subsec:GDSAmodel}. The optimization can be written as 
\begin{equation}
\theta^\star = \underset{\theta}{\operatorname{argmax}}~~ \mathbb{E}_{\tau \sim \pi_\theta(\tau)}\left[\sum_{t} r(s_t, a_t) \right],
\end{equation}
where $r(s_t,a_t)$ is the reward for taking action $a_t$ in state $s_t$, and $\tau=[s_1,a_1,s_2,a_2,\dots,s_T,a_T ]$. We employ policy gradient for optimization using the following gradient:
\begin{equation}
\begin{split}
\frac{1}{N} \sum_{i,t=1}^{N,T}(&\nabla_\theta \log(\pi_\theta(a_{i,t}|s_{i,t}))(\sum_{t'=t}^{T}\gamma^{t'-t} r(s_{i,t'}, a_{i,t'}))),
\end{split}
\label{eq:thetagradfinal}\end{equation}
with a discount factor $\gamma=0.9$, and $N=2$.
We summarize the pseudo code for RL training of the GDSA agent in Algorithm~\ref{alg:rl}.
\begin{small}
\begin{algorithm}[t]
\caption{Training GDSA agent}\label{alg:rl}
\begin{algorithmic}[1]
\State \textbf{Input}: $\mathcal{D} = [(x_1, y_1), (x_2, y_2), \dots]$
\State \textbf{Initialise model parameters}: $\theta$
\For{\emph{epoch\_index} \textbf{in} $[1,2,\dots,\text{num\_epochs}]$}
\State Gradients: $\mathcal{G} = [\,\,]$
\State Sample a random input $x_*$ with its label $y_*$
\State Split $x_*$ into AUs $[x_*^{(1)},x_*^{(2)},\dots]$
\State Get AU rep.: $f=[[\text{g}(x_*^{(1)}), \tau_{1}],[\text{g}(x_*^{(2)}), \tau_{2}],\dots]$
\State Get the category embedding: $\psi_\theta(y_*)$
\For{\emph{play\_index} \textbf{in} $[1,2,\dots,N]$}
\State Candidate-AU: $L_a = f$ 
\State Chosen-AU: $L_b = [\,\,]$
\State Gradient-buff: $G = [\,\,]$
\State Reward-buff: $R = [\,\,]$
\For{\emph{pass\_index} \textbf{in} [1,2,\dots,T]}
\State Get chosen AU embed. using RNN: $\omega_\theta(L_b)$
\State Concatenate feats: $[[\omega,L_a^{(1)},\psi],\dots]$
\State $\pi=\operatorname{softmax}(\phi_\theta([\omega,L_a^{(1)},\psi]),\dots])$
\State Draw an AU $a$ from discrete distribution $\pi$
\State Move the $a$-th AU from $L_a$ to $L_b$
\State Calculate gradient $\nabla_\theta \log \pi[a]$ \& add it to $G$
\State Calculate reward using Eq.~\ref{eq:reward} \& add it to $R$
\EndFor
\State Calculate reweighed gradients using Eq.~\ref{eq:thetagradfinal}
\State Add the \emph{sum} of reweighed gradients to $\mathcal{G}$
\EndFor
\State Do gradient ascent using the \emph{average} of $\mathcal{G}$
\EndFor
\State \textbf{Output}: $\theta$
\end{algorithmic}
\end{algorithm}
\end{small}

\begin{figure}[t]
\centering
\includegraphics[width=\linewidth]{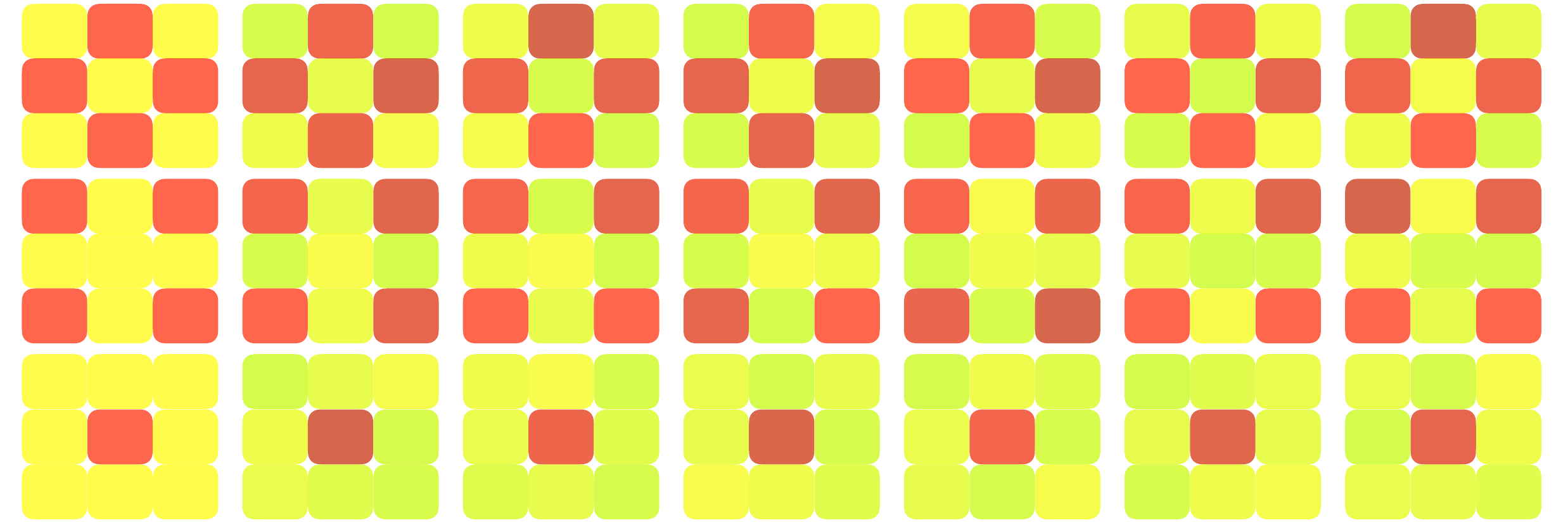}
\caption{Samples in a synthetic dataset. The first column is the class prototype, and the remainders are observed samples.}
\vspace{-0.5cm}
\label{fig:toydata}
\end{figure}

\begin{figure}[t]
\centering
\includegraphics[width=\linewidth]{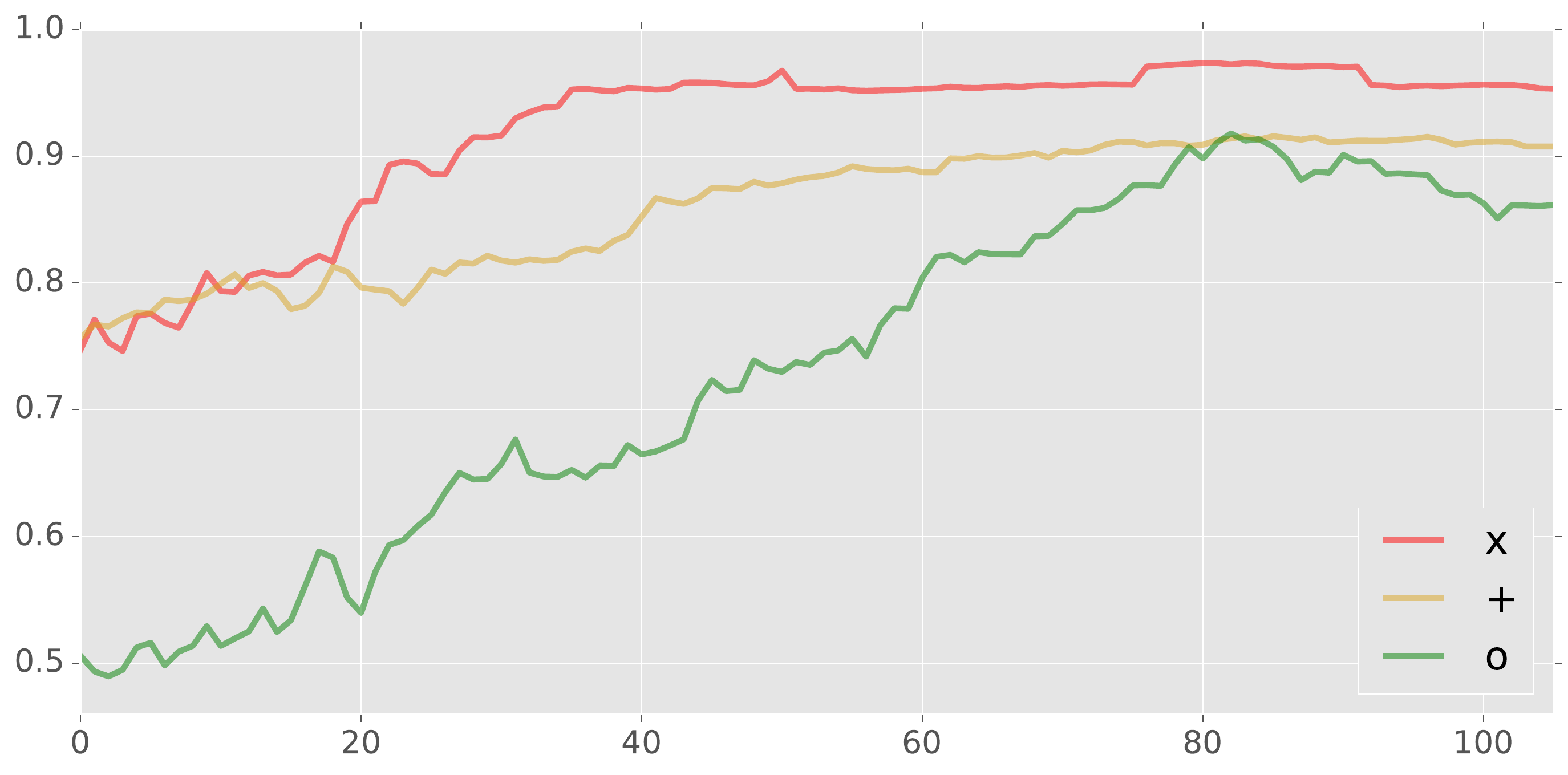}
\caption{Synthetic data GDSA agent. X-axis: train iterations. Y-axis: test accuracy (recognition of 2-pixel image sequences).}
\vspace{-0.5cm}
\label{fig:toyexp}
\end{figure}

\keypoint{A synthetic illustration}
We apply our method to a synthetic example for illustration. We introduce a simple $3\times3$ image format, generated as a sequence of 9 binary samples in raster scan sequential order. Each AU is one pixel, and there are $2^9$ unique image categories.  We choose $3$ classes, corresponding to the first column in Fig.~\ref{fig:toydata}, denoted as `$\times$', `$+$', and `o' respectively. To introduce  intra-class variability, observed samples are perturbed by Gaussian noise. A key observation in Fig.~\ref{fig:toydata} is that, to recognize a category, not all AUs (pixels) are necessary. For example, in a sequence of only two AUs, if one is the corner and other is the centre, then it must be the '$\times$' category. This creates room for simplification and re-ordering of the AU sequence to produce a shorter but information-preserving sequence.

Training the RL agent for this problem, we expect it to pick few AUs that maximize  recognizability. We limit the AU selection budget to 2 (i.e., two pixel output images). As shown in Fig.~\ref{fig:toyexp}, the agent produces output sequences with $\sim90\%$ probability of being correctly classified by a linear classifier. This is is significantly better than its randomly initialized state (a policy that picks two strokes at random), for which the performance is about $50\%\sim70\%$.

\section{Experiments}
\label{sec:Experiments}


\keypoint{Generic implementation details} Our model is implemented in Tensorflow \cite{tensorflow2015-whitepaper}. The RNN used in the GDSA framework to process the chosen AU sequence is implemented with single layer gated recurrent units (GRU) with 128 hidden cells. The GRU output of dimension $1\times128$ is fed to a fully connected layer to get the chosen-strokes embedding of dimension $1\times18$. The candidate AU embedding, obtained by feeding the AU representation (fixed length feature vector concatenated with time-stamp) into a fully connected layer, is of dimension $1\times9$. The class embedding is of dimension $1\times3$. The complete embedding, obtained by concatenating the three previous embeddings and feeding into a fully connected layer, is sized $1 \times 15$. Both the code and trained models will be made public.

\keypoint{Setting discussion} As mentioned earlier, we proposed a new problem setting and associated solution for abstraction learning. While contemporary learning for summarization requires annotated target summaries \cite{berger2013style,cheng2016neural,narayan2018ranking,song2015tvsum,zhang2016video}, we require instead a goal function. The goal function is itself learned from metadata that is often already available  or easier to obtain than expensive gold-standard summaries (e.g., sentiment label for text). Since the goal (task-specific vs.~generic summaries) and data requirements (weak.~vs strongly annotated) of our method are totally different, we cannot compare to conventional summarization methods.

\subsection{Sketch  abstraction}\label{subsec:SketchAbstraction}

\begin{figure*}
	\includegraphics[width=\linewidth]{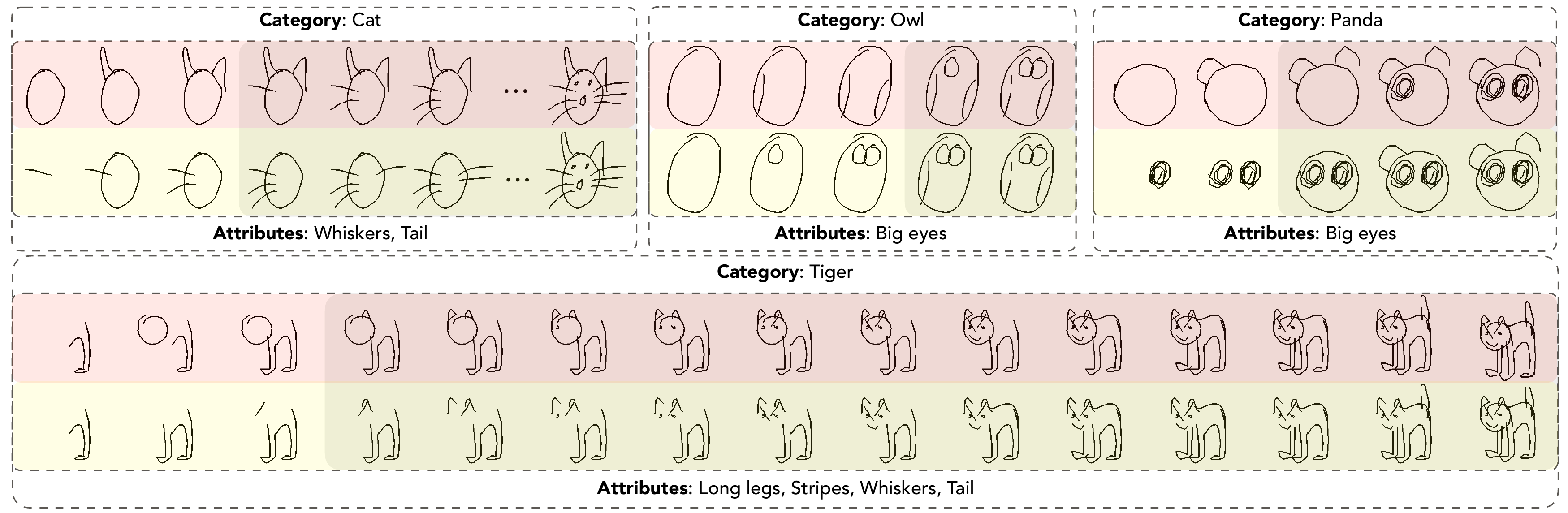}
	\caption{Qualitative comparison of goal-driven stroke sequencing strategies for QuickDraw sketches using the budget size of $25\%$ (light background). In each section: the top row depicts the stroke sequence obtained using GDSA with category-goal, and the bottom row depicts the stroke sequence obtained using GDSA with attribute-goal.}
	\vspace{-0.5cm}
	\label{fig:SketchGoalDrivenQualitativeResults}
\end{figure*}

\keypoint{Dataset} We train our GDSA for sketch using QuickDraw \cite{ha2017neural}, the largest free-hand sketch dataset. As in \cite{muhammad2018learning}, we choose 9 QuickDraw categories namely: \{cat, chair, face, fire-truck, mosquito, owl, pig, purse and shoe\}; using 70,000 sketches per category for training and 5,000 for testing. The average number of strokes in the  9 chosen  categories are \{9.8, 4.9, 6.4, 8.3, 7.2, 9.1, 9.5, 3.6, 3.0\}.

\keypoint{Implementation details} We train our agent with $K=50,000$ episodes, reward scaling factor $b = 100$ and learning rate $\eta=0.0001$.  We set the budget $B$ to $25\%$ and $50\%$ of the (rounded) average number of strokes per category. For reward computation, at each step the list of strokes chosen by our agent is fed into a classifier to determine the probability of the ground truth class. The same is done for the strokes chosen following original input and random order. To this end, we employ two different classifiers: (1) A state-of-the-art convolutional neural network (CNN) - Sketch-a-Net 2.0 \cite{yu2017sketch}, fine-tuned on the 9 QuickDraw categories. The stroke data is rendered as an image before CNN classification.
(2) A three-layer LSTM with 256 hidden cells at each layer, employed in \cite{muhammad2018learning}. It takes an input list of  $(x,y, p)$ (coordinates and pen-state) and feeds its last time-step output to a fully-connected layer with softmax activation which provides the probability distribution for the prediction of the sketch class. After training, this RNN is also used to extract the $256$ dimensional feature vector for each stroke in the candidate stroke pool which is concatenated with one-hot time-stamp vector to get the final AU representation of dimension $D=266$. Note that we do not use Sketch-a-Net for this purpose, due to the sparsity of rendered single strokes images, with which the CNN cannot generate a meaningful representation.

\keypoint{Results} We evaluate the performance of our GDSA model by sketch recognition accuracy when using a budget $B$ of $25\%$ and $50\%$ of the average number of strokes for each category. This evaluation is performed on the testing set of 45,000 sketches. Sketch recognition is achieved using two different classifiers (RNN \cite{muhammad2018learning} and Sketch-a-Net \cite{yu2017sketch}) described previously. We compare our abstraction model with: (1) First $B$ strokes in the original human drawing order. This is a strong baseline, as the data in QuickDraw is obtained by challenging the \textit{player} to draw the object (abstractly) in a time-limited setting - the first few strokes  are thus those deemed important for recognizablity by humans. (2) Random selection of $B$ strokes. (3) DSA \cite{muhammad2018learning}, the state-of-the-art deep sketch abstraction model. Note that to make a fair comparison we adapt \cite{muhammad2018learning} to perform abstraction at stroke level, as the original paper dealt with stroke-segments (five consecutive $(x,y,p)$ elements). (4) DQSN \cite{zhou2018video}, an abstraction model originally proposed for videos. We adapt this model to our setting by plugging in stroke AU representations instead of video-frame features. We also report the performance of the full input sequence without abstraction, which represents the upper bound.
The results  in Table~\ref{tab:SketchAbstraction} show that our GDSA agent outperforms all the other methods. The improvement in performance is most evident for the harder $B=25\%$ budget, confirming the ability of our GDSA model to learn an efficient selection policy. In particular,  both DSA and DQSN are restricted by the original input AU order with a fixed 2-state action space, resulting in sub-optimal selection.

\begin{table}[t]
	\centering
	\resizebox{\linewidth}{!}{%
		\begin{tabular}{l|cc|cc}
			\hline
			\multicolumn{1}{l}{Budget }  &   \multicolumn{2}{c}{25\%} & \multicolumn{2}{c}{50\%}  \\ \hline
			Method & RNN & Sketch-a-Net & RNN & Sketch-a-Net  \\ \hline 
			Human & 36.66 & 62.08 &  66.73 & 75.90 \\
            Random & 22.67 & 41.06 & 45.65 & 65.47 \\
            DSA \cite{muhammad2018learning} & 38.36 & 65.05 & 67.89 & 81.50 \\
            DQSN \cite{zhou2018video} & 38.11 & 64.58 & 67.50 & 80.31 \\
			GDSA & \textbf{50.50} & \textbf{71.92} & \textbf{71.75} & \textbf{86.15} \\ \hline
			Upper bound & 87.77 & 91.99 & 87.77 & 91.99 \\  \hline
		\end{tabular}%
	}
	\caption{Category recognition (acc. \%) of the abstracted sketches.}
	\vspace{-0.5cm}
	\label{tab:SketchAbstraction}
\end{table}

\begin{figure*}
	\includegraphics[width=\linewidth]{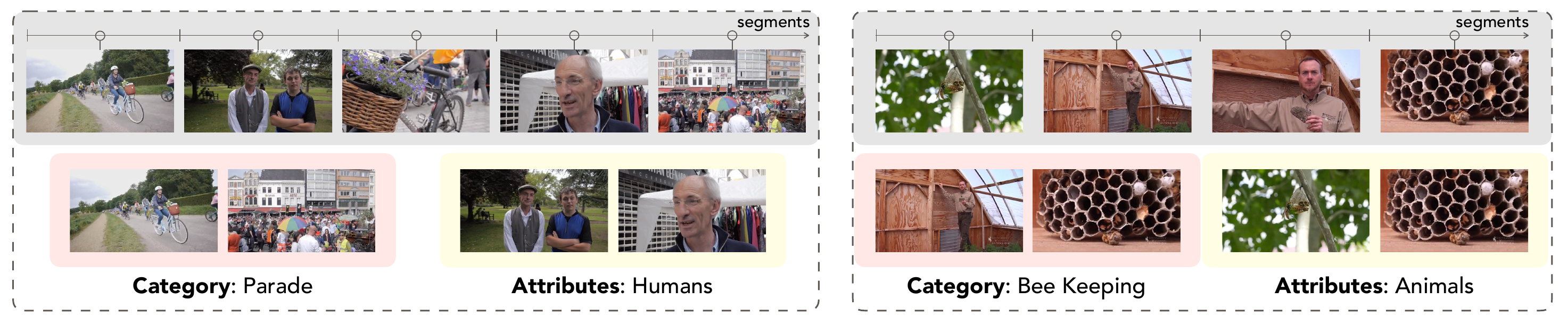}
	\caption{Qualitative comparison of goal-driven stroke sequencing strategies for TVSum videos using the budget size of $25\%$. Grey: Full video clip. Pink: GDSA with category preservation goal. Yellow: GDSA with attribute preservation goal.}
	\vspace{-0.5cm}
	\label{fig:VideoGoalDrivenQualitativeResults}
\end{figure*} 

\keypoint{Abstraction with a different goal} A key feature of our approach is the ability to select different input properties that should be preserved during the abstraction. In this experiment, we demonstrate this capability by contrasting attribute preservation with the category-preservation. We do this by selecting 9 animal categories from QuickDraw  (cat, mouse, owl, panda, pig, rabbit, squirrel, tiger and zebra) and defining 5 animal attributes:
whiskers (cat, mouse, rabbit, tiger), tail (cat, mouse, pig, rabbit, squirrel, tiger, zebra), stripes (tiger, zebra), long-legs (tiger, zebra), big-eyes (owl, panda). We train two separate Sketch-a-Net 2.0 models to recognize the above mentioned categories and attributes. These are then plugged in the reward generator to train GDSA, with budget $B=25\%$.
Qualitative comparison of the results of category vs attribute preservation are shown in Fig.~\ref{fig:SketchGoalDrivenQualitativeResults}. We can see clearly that changing the goal has a direct impact on the abstraction strategy. E.g., preserving the salient cat category cue (ears) vs.~the requested attribute (whiskers).

\subsection{Video abstraction}
\label{subsec:VideoAbstraction}
\keypoint{Dataset} We train GDSA for video using the TVSum dataset \cite{song2015tvsum}, with the primary objective of preserving video category information. This dataset contains 10 categories: \{changing vehicle tire, getting vehicle unstuck, groom animal, making sandwich, parkour, parade, flash mob gathering, bee keeping, bike tricks, and dog show\}. We use 40 video samples for training, and 10 for testing. The video length vary from 2 to 10 minutes. Following the common practice \cite{zhang2016video, zhou2018video} we down-sample videos to 1 fps, and then use shot-change data to merge 5 consecutive shots to form coarse segments. After this, the average number of segments per video in each category are \{13.5, 8.9, 10.4, 9.3, 7.8, 11.0, 10.4, 10.5, 12.1, 9.9\}  respectively.

\keypoint{Implementation details} We train our agent with $K=100$ episodes, reward scaling factor $b = 100$ and  learning rate $\eta=0.0001$. We set the budget $B$ to $25\%$ and $50\%$ of the average number of segments for each category. Additionally, we test a budget of one segment to find the single most relevant segment in each video. For reward computation we use a multi-class bidrectional GRU classifier, originally proposed in \cite{zhou2018video}. This classifier, after training, is also used to extract the fixed dimension (512) feature vector which is concatenated with the time-stamp vector to obtain the final AU representation ($D=522$) for each segment in the candidate pool.

\keypoint{Results} The performance of GDSA is evaluated by category recognition accuracy, at three budget values of $1$ segment, $25\%$ and $50\%$ of the average number of segments per category. Following \cite{zhou2018video} this evaluation is performed by doing 5-fold cross-validation. Category recognition is performed using the aforementioned classifier. We compare with: (1) First $B$ segments in the original order. (2) Random $B$ segments. (3) DSA \cite{muhammad2018learning}, adapted to videos by substituting stroke AU vector with video-segment AU. (4) The state-of-the-art DQSN \cite{zhou2018video}, which is adapted to be trained with a category-recognition based reward for fair comparison. We also compute the upper bound of the input  video without abstraction.  The results in Table~\ref{tab:VideoAbstraction} show that our GDSA agent outperforms all competitors by significant margins. 
\begin{table}[t]
	\centering
	\resizebox{0.7\linewidth}{!}{%
		\begin{tabular}{l|c|c|c}
			\hline 
			\multicolumn{1}{l}{Budget }  & \multicolumn{1}{c}{1 segment} &  \multicolumn{1}{c}{25\%} & \multicolumn{1}{c}{50\%}  \\ \hline 
			Method & RNN & RNN & RNN \\ \hline 
			Original & 28.0 & 28.0 & 28.0  \\
            Random & 28.0 & 32.0 & 34.0  \\
            DSA \cite{muhammad2018learning} & 42.0 & 62.0 & 72.0 \\
            DQSN \cite{zhou2018video} & 44.0 & 64.0 & 72.0 \\
			GDSA model & \textbf{68.0} & \textbf{74.0} & \textbf{76.0} \\ \hline 
			Upper bound & 78.0 & 78.0 & 78.0 \\  \hline
		\end{tabular}%
	}
	\caption{Category recognition (accuracy \%) of video samples.}
	\vspace{-0.5cm}
	\label{tab:VideoAbstraction}
\end{table}

\keypoint{Abstraction with a different goal} In order to demonstrate the goal-driven abstraction capability of our model, we first define 5 category level attributes: animals (dog show, grooming animals, bee keeping), humans (parkour, flash mob gathering, parade), vehicles (changing vehicle tire, getting vehicle unstuck), food (making sandwich), bicycle (attempting bike tricks)\}. Using the same classifier architecture used for category classification, we train the attribute classifier. This is then plugged in the reward function to guide training, with $B=25\%$. Some qualitative results are shown in Fig.~\ref{fig:VideoGoalDrivenQualitativeResults}. We can clearly observe that abstracted output varies according to the goal function. E.g., preserving parade related segments (category) vs.~segments depicting humans (attribute).

\begin{figure*}
    \centering
	\includegraphics[width=\linewidth]{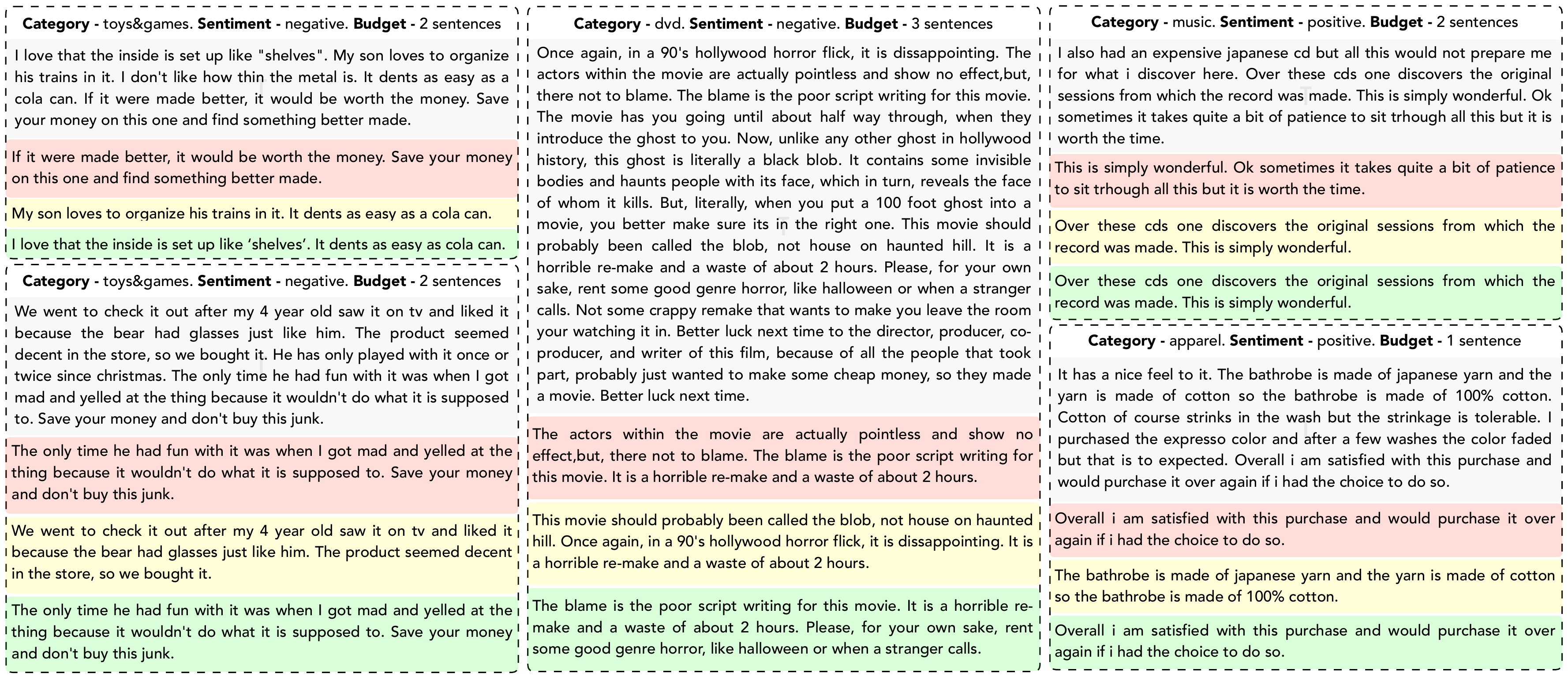}
	\caption{Qualitative comparison of goal-driven summarization for Amazon product reviews with budget $25\%$. Grey: Full review. Pink: GDSA for sentiment. Yellow: GDSA for category. Green: GDSA for helpfulness.}
	\vspace{-0.5cm}
	\label{fig:GoalTextQualitativeResults}
\end{figure*}

\subsection{Text abstraction}
\label{subsec:TextAbstraction}


\keypoint{Dataset} We train our GDSA model for text using the Amazon Review dataset \cite{mcauley2015image}. We aim to preserve positive/negative review sentiment (1-2 stars as negative, 4-5 stars as positive). We choose 9 categories: \{apparel, books, dvd, electronics, kitchen and houseware, music, sports and outdoors, toys and games, and video\}, based on the availability of equal number of positive and negative reviews. The average number of sentences per category are \{3.5, 8.2, 8.9, 5.6, 4.8, 6.8, 5.4, 4.9, 7.7\} respectively. We use 1400 reviews per category for training and 600 for testing.

\keypoint{Implementation details} We train our agent with $K=10000$ episodes, reward scaling factor $b = 100$ and  learning rate $\eta=0.0001$. We set the budget $B$ to $25\%$ and $50\%$ of the average number of sentences for each category. Additionally, we have a budget of one sentence to find the single most relevant sentence in each review. We use two different sentiment classifiers, both using Glove embedding \cite{pennington2014glove} to represent each word as fixed dimension vector: (1) A state-of-the-art hierarchical attention network (HAN) \cite{yang2016hierarchical} for text classification, trained for binary sentiment analysis on the 9 review categories. (2) A RNN built with a single layer LSTM of 64 hidden cells. It takes as input the list of word embeddings and feeds its last time-step output to a fully-connected layer with softmax activation to predict the sentiment. These classifiers, once trained, are also used to extract a fixed dimension (256/64) feature which is concatenated with the time-stamp vector to get the final AU representation ($D=266/74$) for each sentence in the candidate sentence pool, for the respective GDSA models.

\keypoint{Results} We evaluate the performance of our GDSA model by sentiment recognition accuracy with three budgets of $1$ sentence, $25\%$ and $50\%$ of the average number of sentences per category. This evaluation is performed on the testing set of 5,400 reviews. Sentiment recognition uses the two classifiers (RNN and HAN \cite{yang2016hierarchical}) described above. We compare with: (1) First $B$ sentences in the original order. (2) Random $B$ sentences. (3) DSA    \cite{muhammad2018learning} and (4) DQSN \cite{zhou2018video}, both adapted to text by plugging in the sentence AU representation instead of stroke and frame AU representation. Upper bound represents the performance of the full review without abstraction. The results in Table~\ref{tab:TextAbstraction} show that our GDSA agent again outperforms all competitors. 

\begin{table}[t]
	\centering
	\resizebox{\linewidth}{!}{%
		\begin{tabular}{l|cc|cc|cc}
			\hline 
			\multicolumn{1}{l}{Budget }  & \multicolumn{2}{c}{1 sentence} &  \multicolumn{2}{c}{25\%} & \multicolumn{2}{c}{50\%}  \\ \hline 
			Method & RNN & HAN & RNN & HAN & RNN & HAN \\ \hline 
			Original & 59.70 & 67.47 & 66.57 & 76.06 & 70.73 & 80.27 \\
            Random & 61.16 & 69.04 & 66.44 & 77.14 & 70.98 & 81.57 \\
            DSA \cite{muhammad2018learning} & 66.37 & 72.42 & 71.58 & 80.02 & 73.36 & 83.47 \\
            DQSN \cite{zhou2018video} & 65.70 & 71.40 & 71.93 & 80.25 & 73.20 & 83.77 \\
			GDSA model & \textbf{70.64} & \textbf{83.77} & \textbf{73.39} & \textbf{86.08}  & \textbf{74.11} & \textbf{86.12}  \\ \hline 
			Upper bound & 76.41 & 86.66 & 76.41 & 86.66 & 76.41 & 86.66 \\  \hline
		\end{tabular}%
	}
	\caption{Sentiment recognition (accuracy \%) of review summaries.}
	\vspace{-0.5cm}
	\label{tab:TextAbstraction}
\end{table}

\keypoint{Abstraction with different goals} We next demonstrate the GDSA model's goal-driven summarization capability by training instead to preserve (1) product-category (multi-class), and (2) helpfulness (binary) data. HAN classifier and $B=25\%$ are used. 
Some qualitative results are shown in Fig.~\ref{fig:GoalTextQualitativeResults}. We can observe that depending on the abstraction goal, the output varies to preserve the information relevant to the goal. 


\section{Conclusion}
\label{sec:Conclusion}

We have introduced a new problem setting and effective framework for goal-driven sequential data abstraction. It is driven by a goal-function, rather than needing expensively annotated ground-truth labels, and also uniquely allows selection of the information to be preserved rather than producing a single general-purpose summary. Our GDSA model provides improved performance in this novel abstraction task compared to several alternatives. Our reduced data requirements, and new  goal-conditional abstraction ability enable different practical summarization applications compared to those common today. 

{\small
\bibliographystyle{ieee_fullname}

}

\end{document}